# Design of Cavity Backed Slotted Antenna using Machine Learning Regression Model

Vijay Kumar Sutrakar
*ADE, DRDO*
Bangalore, India
vks.ade@gov.in

Anjana P K
*CE, ADE, DRDO*
Bangalore, India
pkanjana505@gmail.com

Rohit Bisariya
*CE, ADE, DRDO*
Bangalore, India
rohit881900@gmail.com

Soumya K K
*CE, ADE, DRDO*
Bangalore, India
soumyasajil96@gmail.com

Gopal Chawan M
*ADE, DRDO*
Bangalore, India
gopalchawan95@gmail.com

*Abstract*—In this paper, a regression-based machine learning model is used for the design of cavity backed slotted antenna. This type of antenna is commonly used in military and aviation communication systems. Initial reflection coefficient data of cavity backed slotted antenna is generated using electromagnetic solver. These reflection coefficient data is then used as input for training regression-based machine learning model. The model is trained to predict the dimensions of cavity backed slotted antenna based on the input reflection coefficient for a wide frequency band varying from 1 GHz to 8 GHz. This approach allows for rapid prediction of optimal antenna configurations, reducing the need for repeated physical testing and manual adjustments, may lead to significant amount of design and development cost saving. The proposed model also demonstrates its versatility in predicting multi frequency resonance across 1 GHz to 8 GHz. Also, the proposed approach demonstrates the potential for leveraging ma- chine learning in advanced antenna design, enhancing efficiency and accuracy in practical applications such as radar, military identification systems and secure communication networks.

*Index Terms*—Cavity Backed Slotted Antenna (CBSA), Machine Learning, Regression Model, Reflection Coefficient (RC), Voltage Standing Wave Ratio (VSWR)

## I. INTRODUCTION

Cavity Backed Slotted Antenna systems have been an integral part of military and civil aviation technology since their inception during World War II. These systems enable quick and reliable identification of friendly units, reducing the risk of friendly fire and ensuring mission success in hostile or congested environments. The CBSA system operates by transmitting interrogation signals from ground-based or airborne radars and receiving response signals from transponders located on friendly aircraft or vehicles. A key challenge for CBSA systems is maintaining robust communication links under various operational conditions, particularly in the L-band frequency range (1–2 GHz), which is widely adopted for its ability to penetrate obstacles and provide reliable performance in adverse weather conditions [1]. The design of CBSA, therefore, plays a crucial role in ensuring that these communication systems function optimally, providing both range and accuracy, while minimizing signal interference. CBSA have also been widely used for radar, satellite, and wireless communication applications [2].

Traditionally, the design of CBSA has relied heavily on manual methods, involving numerous simulation cycles and physical prototyping to fine-tune the antenna's structure. Parameters such as slot length, slot width and slot orientation must be optimized to achieve the desired radiation charac- teristics and minimize reflection losses, which can degrade the antenna's performance. This trial-and-error approach is not only time-consuming but also costly, particularly when precise and highly efficient designs are required for mission-critical applications [3]. Furthermore, with increasing demands for antennas that can operate in more complex environments such as urban warfare, jamming-prone battlefields or congested civilian air spaces, there is a growing need for more advanced and efficient design methods.

In the last many years, a large number of surrogate models have been used for the design and optimization of antenna systems. It helps to reduce the computational burden of using full wave solvers [4-6]. However, usage of machine learning (ML) methods are promising for further reduction in computational and design cycle cost of antennas. Differ- ent machine learning techniques such as Gaussian process regression (GPR) [4,5] and artificial neural network (ANN) [6] have been used in the recent past. ML methods have also been used globally for various antenna design applications [7-13], including antenna design for communication applications [7,8], wireless applications [10] etc.

In recent years, ML has emerged as a powerful solution to tackle complex optimization problems across a wide range of engineering fields. ML models, which excel at identifying patterns in data and making predictions, have begun to revolutionize traditional design processes by offering more efficient ways to optimize parameters. In antenna design, ML technique can be used to predict the optimal configurations of an antenna based on performance data, such as reflection coefficients, thereby minimizing the need for exhaustive simulations and manual adjustments [14]. This is especially relevant in the design of CBSA, where small variations in physical parameters can have significant impacts on the antenna's overall performance.

The integration of machine learning into antenna design has opened new avenues for innovation, allowing for faster and more accurate predictions of antenna performance. By training machine learning models on datasets generated from previous designs, engineers can now explore a wide range of parameter configurations without the need for iterative physical testing.

In this paper, a machine learning-based approach is proposed to streamline the design process of CBSA. Specifically, key physical parameters, i.e., Slot Length, Slot Width and Slot Angle of CBSA is predicted using machine learning model. Further details of modelling and simulations are provided in Section II. Results and discussions are provided in Section III, followed by concluding remarks in Section IV.

## II. MODELLING AND SIMULATION DETAILS

### A. Electromagnetic modelling and simulation details of antenna

A CBSA with perfect electrical conducting (PEC) material properties is considered in the present study for generating the reflection coefficients. The following parameters are initially considered for the antenna simulation: Width of antenna, W = 120 mm, length of antenna, L = 140 mm, thickness of antenna radiating surface, T = 2.5 mm, cavity height, H = 16.3 mm, slot dimensions, $Sl_1$ = 127.5 mm, $Sl_2$ = 49 mm, $Sw_1$ = 6 mm, and $Sw_2$ = 6 mm, as shown in Figures 1(a) and 1(b). The slot is shown at angle $\vartheta$ is 0 degree in Figure 1. The coaxial feed with inner diameter, d = 2.65 mm and outer diameter, D = 9.91 mm is considered in the present study. The coaxial feed is considered of glass PTFE with relative permittivity is 2.5. The L, W, T, H, $Sl_2$, and $Sw_2$ is kept constant in the present study. Only $Sl_1$, $Sw_1$, and $\vartheta$ is varied for generating the reflection coefficient data set (further details are provided in Table 1). The electromagnetic simulations are carried out using Ansys HFSS [15]. The reflection co-efficient and VSWR performance of the CBSA with the above given dimensions are shown in Figure 2. The total gain obtained for the CBSA with above mentioned dimensions is 5.99 dB at 1.05 GHz for both $\phi$ = 0 and 90; 1.02 dB at Φ = 90 and 8.29 dB at Φ = 0 at 4.35 GHz; 9.40 dB at Φ = 90 and 12.10 dB at Φ = 0 at 7.47 GHz, as shown in Figure 3.

### B. Machine Learning Model details

Total 3778 samples are considered in the present study (refer Table – 1 for further details). The input to the ML model consists of reflection coefficients with varying frequencies. The model predicts Only $Sl_1$, $Sw_1$, and $\vartheta$ as output. In the present work, Lasso regression which is a regularized version of linear regression is used. The raw input data has a dimension of (3778, 1001) representing 3778 different antenna configurations with 1001 frequency points varying from 1 GHz to 8 GHz with step of 10 MHz. After completion of pre-processing steps, the data is converted into dimensions of (3778, 11476). These pre-processing steps includes removal of outliers, standardization, applying Principal Component Analysis (PCA), and Polynomial Feature Expansion. Further details of each of these features are provided subsequently.

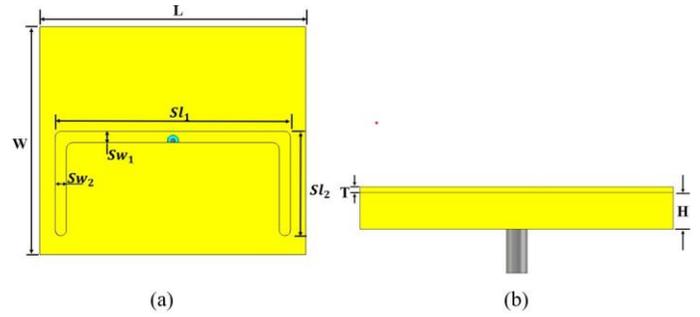

Fig. 1. (a) Top view and (b) Side view of CBSA

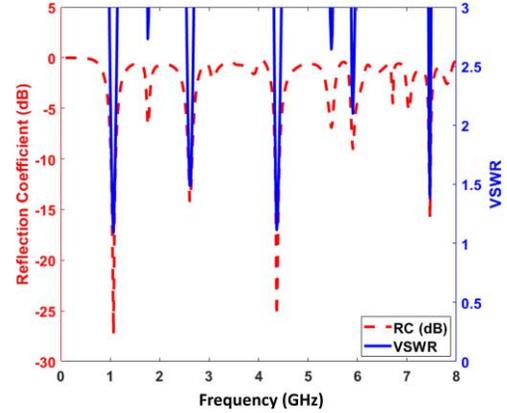

Fig. 2. Reflection Coefficient (RC) and Voltage Standing Wave Ratio (VSWR) of CBSA.

Outliers can significantly distort the learning process of machine learning models, especially when dealing with complex physical phenomena like antenna behavior. Outliers may occur due to measurement errors, noise in the data, or extreme configurations of the antenna that lead to anomalous performance. The removal of outliers ensures that the model is trained on reliable data and that the predictions are not skewed by these anomalous values. In this work, Interquartile Range (IQR) have been applied to removal outliers. This method helps to identify data points that are far from the central tendency (mean or median) and might negatively influence the model's performance. After the removal of outliers, the dataset retains only those configurations that are within a statistically acceptable range of s-parameter values, which represent realistic antenna behavior [16].

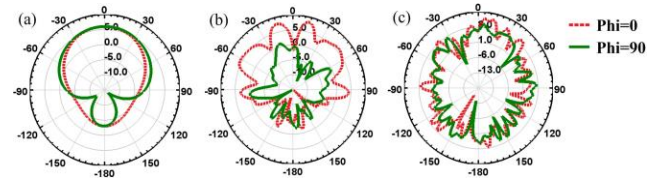

Fig. 3. 2D Radiation patterns of CBSA at (a) 1.05 GHz, (b) 4.35 GHz and (c) 7.47 GHz

TABLE I
PARAMETER VARIATION OF CBSA ANTENNA TAKEN FOR THE DATASET

| Angle (degree) | $Sl_1$ (mm) | | | $Sw_1$ (mm) | | | Available Samples |
|---|---|---|---|---|---|---|---|
| | min | max | Step Size | min | max | Step Size | |
| 0 | 30 | 130 | 3 | 5 | 30 | 2 | 476 |
| 10 | 25 | 125 | 3 | 5 | 45 | 3 | 476 |
| 20 | 25 | 120 | 3 | 5 | 37 | 3 | 396 |
| 30 | 25 | 120 | 3 | 5 | 35 | 3 | 363 |
| 40 | 25 | 110 | 3 | 5 | 33 | 3 | 290 |
| 50 | 25 | 110 | 3 | 5 | 30 | 3 | 261 |
| 60 | 25 | 100 | 3 | 5 | 33 | 3 | 260 |
| 70 | 25 | 100 | 3 | 5 | 35 | 3 | 286 |
| 80 | 25 | 100 | 3 | 5 | 60 | 3 | 494 |
| 90 | 30 | 130 | 3 | 5 | 30 | 2 | 476 |

After removing outliers, standardization is applied to the input data to normalize the scale of the features. By transforming all features to a common scale, the machine learning model can treat all input features equally, improving its ability to learn from the data. Standardization is particularly important for models that rely on distance metrics, such as (PCA) and models that assume normal distributions [17].

Given that the original dataset contains 1001 features, dimensionality reduction is essential to mitigate the risk of overfitting and to reduce computational complexity. PCA reduces the dimensionality while retaining the critical information needed for predicting antenna parameters [18]. In this case, PCA is applied after standardization and it likely reduces the input data to a lower-dimensional space that captures the most significant patterns in the s-parameter data. For this study, PCA helps to filter out the noise and redundancy in the data, providing the machine learning model with a more concise and informative input. After applying PCA, the dataset remains suitable for further processing without losing key information required for predicting the output of the model.

Next, Polynomial Feature Expansion is used for capturing relationship between the reflection coefficient and the antenna parameters (i.e. $Sl_1$, $Sw_1$, and $\vartheta$) [19]. This technique gener- ates new features by computing polynomial combinations of the original features, thereby allowing the model to account for interactions between features and capture higher-order relationships that may not be visible in the original feature space and so on. This expansion significantly increases the number of features, allowing the machine learning model to better approximate the complex, nonlinear behavior of the antenna. In this case, the dimensionality of the input data increases from the PCA-transformed space 150 to 11476 features after polynomial expansion

After applying these pre-processing steps, i.e., outlier removal, standardization, PCA and polynomial feature generation, the input to the machine learning model has a final dimensionality of (3778, 11476). This expanded feature set contains both the original features and their polynomial interactions, providing the model with a rich dataset. In this work, Least Absolute Shrinkage and Selection Operator (Lasso) regression is used with alpha ($\alpha$) = 0.01, here, alpha is hyper parameter [20].

## III. RESULTS AND DISCUSSIONS

The Lasso Regression model was evaluated on both training and test datasets, using two performance metrics, i.e. R2 score and Mean Squared Error (MSE). These metrics offer a comprehensive view of how well the model captures the relationship between the reflection coefficient and the slot dimensions, i.e. $Sl_1$, $Sw_1$, and $\vartheta$. The dataset is split into training (80%) and testing (20%). The model achieved an R2 score of 0.9989 and MSE of 0.5375 for the training data set. The test data predicts R2 score of 0.9984 and 0.7955 for MSE. The performance of the Lasso Regression model was further validated by comparing the predicted slot dimensions, i.e. $Sl_1$, $Sw_1$, and $\vartheta$ against their true values, as shown in Table 2 for five random test samples. Subsequently, those predicted slot dimensions (refer Table 2) are modelled in Ansys HFSS [15] and reflection parameters are generated. The comparison of reflection coefficient for the true verses predicted slot dimensions are shown in Figure 4. A very close match is obtained between the true and predicted values.

TABLE II
TRUE AND PREDICTED VALUES OF $Sl_1$, $Sw_1$, AND $\vartheta$ FOR FIVE RANDOM TEST DATA

| Sl No | True | | | Predicted | | | Percentage Error (%) | | |
|---|---|---|---|---|---|---|---|---|---|
| | $Sl_1$ | $Sw_1$ | $\vartheta$ | $Sl_1$ | $Sw_1$ | $\vartheta$ | $Sl_1$ | $Sw_1$ | $\vartheta$ |
| (a) | 109 | 5 | 30 | 108 | 6 | 29 | 0.91 | 16.6 | 3.3 |
| (b) | 52 | 14 | 70 | 51 | 14 | 70 | 1.92 | 0 | 0 |
| (c) | 67 | 14 | 30 | 66 | 14 | 30 | 1.49 | 0 | 0 |
| (d) | 126 | 9 | 10 | 126 | 9 | 11 | 0 | 0 | 10 |
| (e) | 34 | 35 | 40 | 32 | 35 | 39 | 5.88 | 0 | 2.5 |

In order to further evaluate the generalization capability of the model; two different cases are considered. In case – 1, $Sl_1$ and $Sw_1$ are varied for a given $\vartheta$. In the case – 2, all the parameters, i.e. $Sl_1$, $Sw_1$, and $\vartheta$ are varied. In both the cases, five new random samples are considered. These samples are neither part of test or train data set.

In the case – 1, the samples consist of varying slot lengths and slot widths for a given fixed slot angle of 45 degrees, as shown in Table 3. The proposed model is able to predict the slot dimensions very accurately, as shown in Table 3. Subsequently, these dimensions are used for the antenna simulations in Ansys HFSS [15]. The comparison of reflection coefficient for the true and predicted values for the five random samples (with varying $Sl_1$, $Sw_1$, and fixed $\vartheta$= 45 degree) are shown in Figure 5. It can be seen from the Figure 5 that the reflection coefficients obtained using both the true and predicted dimensions exhibit an excellent match. This alignment reinforces the accuracy of the Lasso Regression model, confirming that the predicted slot dimensions are sufficiently precise for practical use in antenna design.

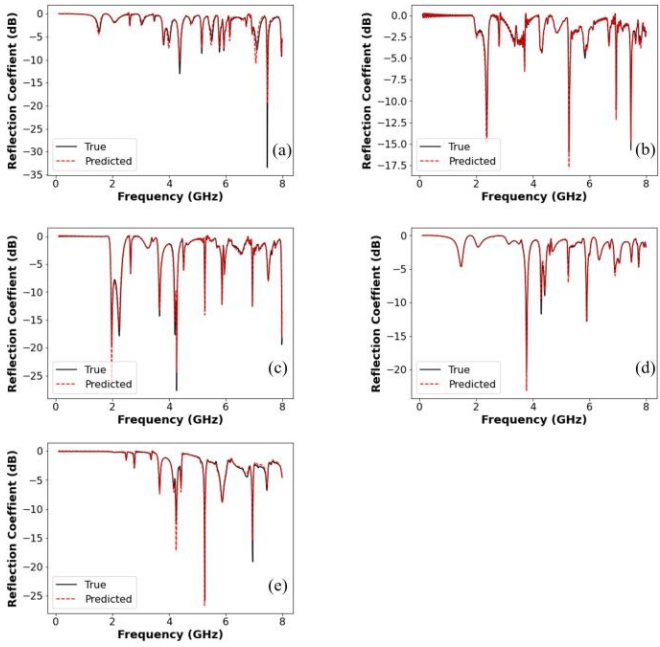

Fig. 4. Simulated (predicted) reflection coefficient curves vs input (true) reflection coefficient for five random test data (refer Table 2 for further details)

TABLE III
TRUE AND PREDICTED VALUES OF $Sl_1$, $Sw_1$, AND FIXED $\vartheta$ OF 45 DEGREE

| Sl No | True | | | Predicted | | | Percentage Error (%) | | |
|---|---|---|---|---|---|---|---|---|---|
| | $Sl_1$ | $Sw_1$ | $\vartheta$ | $Sl_1$ | $Sw_1$ | $\vartheta$ | $Sl_1$ | $Sw_1$ | $\vartheta$ |
| (a) | 80 | 40 | 45 | 78 | 36 | 44 | 2.5 | 1 | 2 |
| (b) | 30 | 37 | 45 | 35 | 37 | 48 | 16 | 0 | 6.6 |
| (c) | 48 | 17 | 45 | 46 | 18 | 44 | 4.1 | 5.8 | 2 |
| (d) | 62 | 11 | 45 | 63 | 9 | 47 | 1.6 | 18 | 4.4 |
| (e) | 68 | 37 | 45 | 68 | 34 | 45 | 0 | 8 | 0 |

TABLE IV
TRUE AND PREDICTED VALUES FOR DIFFERENT ANGLES

| Sl No | True | | | Predicted | | | Percentage Error (%) | | |
|---|---|---|---|---|---|---|---|---|---|
| | $Sl_1$ | $Sw_1$ | $\vartheta$ | $Sl_1$ | $Sw_1$ | $\vartheta$ | $Sl_1$ | $Sw_1$ | $\vartheta$ |
| (a) | 40 | 7 | 77 | 41 | 7 | 78 | 2.5 | 0 | 1.29 |
| (b) | 40 | 5 | 86 | 44 | 5 | 86 | 10 | 0 | 0 |
| (c) | 35 | 16 | 79 | 36 | 14 | 78 | 2.85 | 12.5 | 1.26 |
| (d) | 43 | 8 | 67 | 44 | 8 | 66 | 2.32 | 0 | 1.49 |
| (e) | 38 | 6 | 87 | 44 | 6 | 89 | 15.78 | 0 | 2.29 |

In the case – 2, the samples consist of varying $Sl_1$, $Sw_1$, and $\vartheta$ shown in Table 4. The proposed model is able to predict $Sl_1$, $Sw_1$, and fixed $\vartheta$ very accurately, as shown in Table 4. Subsequently, these dimensions are used for the antenna simulations in Ansys HFSS [15]. The comparison of reflection coefficient for the true and predicted values for the five random samples (with varying $Sl_1$, $Sw_1$, and fixed

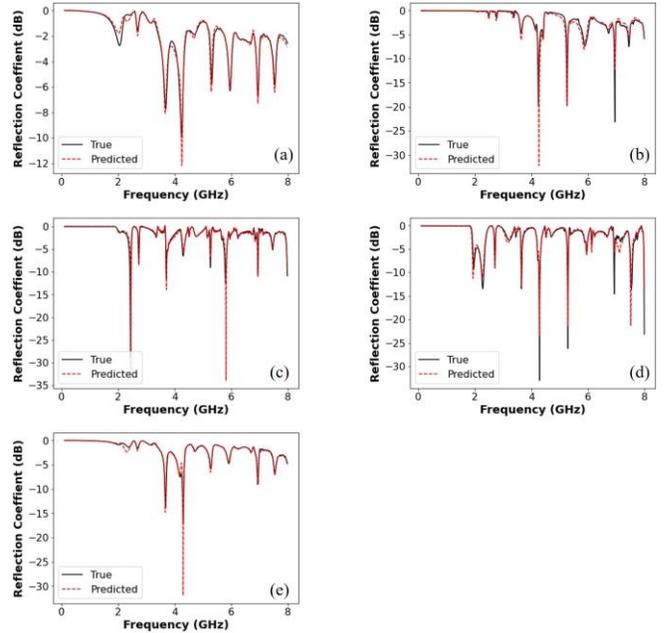

Fig. 5. Simulated (predicted) reflection coefficient curves vs input (true) reflection coefficient for five random data (refer Table 3 for further details)

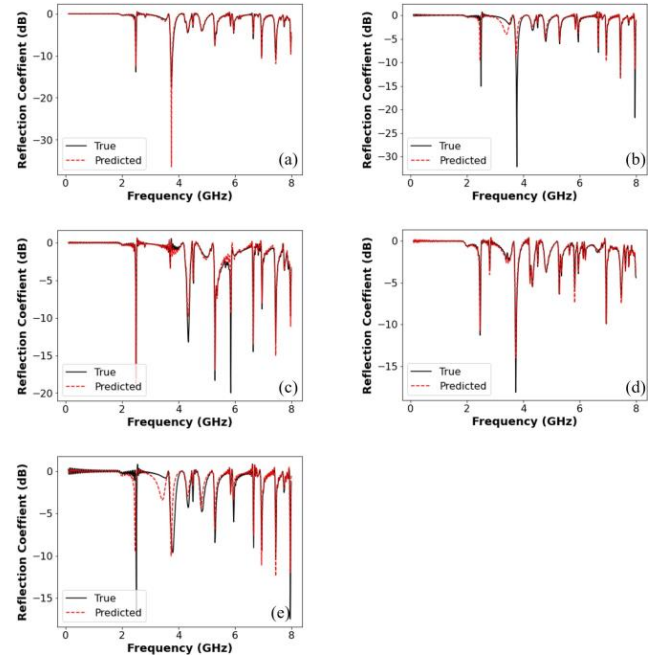

Fig. 6. Simulated (predicted) reflection coefficient curves vs input (true) reflection coefficient for five random data (refer Table 4 for further details)

ϑ) are shown in Figure 6. It can be seen from the Figure 6 that the reflection coefficients obtained using both the true and predicted dimensions exhibit an excellent match.

TABLE V
DIFFERENT RESONANCE FREQUENCY AND THEIR BANDWIDTH AS INPUT FOR PREDICTION OF SLOT DIMENSIONS OF ANTENNA

| Sl No | Center frequency (GHz) | Upper frequency (GHz) | Lower frequency (GHz) | Bandwidth (GHz) |
|---|---|---|---|---|
| (a) | 2.493 | 2.501 | 2.485 | 0.016 |
| (b) | 3.757 | 3.789 | 3.726 | 0.063 |
| (c) | 6.825 | 6.936 | 6.816 | 0.12 |
| (d) | 5.266 | 5.290 | 5.250 | 0.04 |
|  | 6.680 | 6.7123 | 6.651 | 0.0553 |
| (e) | 2.57 | 2.541 | 2.501 | 0.04 |
|  | 5.298 | 5.329 | 5.274 | 0.055 |
|  | 6.846 | 6.870 | 6.815 | 0.0551 |
| (f) | 2.501 | 2.517 | 2.477 | 0.04 |
|  | 5.867 | 5.867 | 5.843 | 0.07 |
|  | 6.649 | 6.649 | 6.625 | 0.055 |
|  | 7.462 | 7.462 | 7.447 | 0.031 |

The application of the Lasso Regression model is further extended for predicting slot dimensions for specific ideal cases where the reflection coefficient is expected to resonate at individual frequencies such as S-Band and C-Band. The model was also tested for combinations of two and three different frequencies to explore its versatility in predicting multi-frequency resonance behavior, as shown in Table 5 (refer Sr. No. (d) for dual frequencies, Sr. No. (e) for triple frequencies and Sr. No. (f) for four different frequencies).

The model successfully predicted slot dimensions for all these cases, achieving accurate results across both single and multiple frequency scenarios. The reflection coefficients obtained from Ansys HFSS [15], using the predicted slot dimensions, matched the expected resonating frequencies as shown in Figure 7. This includes cases where the antenna was designed to resonate at combinations of two or more frequencies. This performance highlights the versatility of the model in handling complex scenarios, such as multi-frequency resonance, where accurate prediction of slot dimensions is critical to ensuring the desired antenna performance. By reliably predicting slot dimensions for both individual and combined resonant frequencies, the Lasso Regression model demonstrates its utility in designing antennas for a wide variety of operating conditions, enabling faster and more efficient design iterations.

## IV. CONCLUSIONS

The proposed regression-based machine learning approach for designing CBSA demonstrates its potential to efficiently predict optimal antenna dimensions based on reflection coefficient data across a wide frequency range (i.e. 1 GHz to 8 GHz). The proposed model is able to predict the antenna dimensions, i.e. $Sl_1$, $Sw_1$, and $ϑ$ accurately. Also, the combinations of two, three, and four different frequencies to explore its versatility in predicting multi-frequency resonance behavior is also demonstrated by the proposed ML model. This method significantly reduces the need for extensive physical testing and manual adjustments, leading to substantial cost savings in design and development. Additionally, it highlights the capability of machine learning to enhance accuracy and efficiency in advanced antenna designs, with practical applications in radar, military identification systems and secure communication networks.

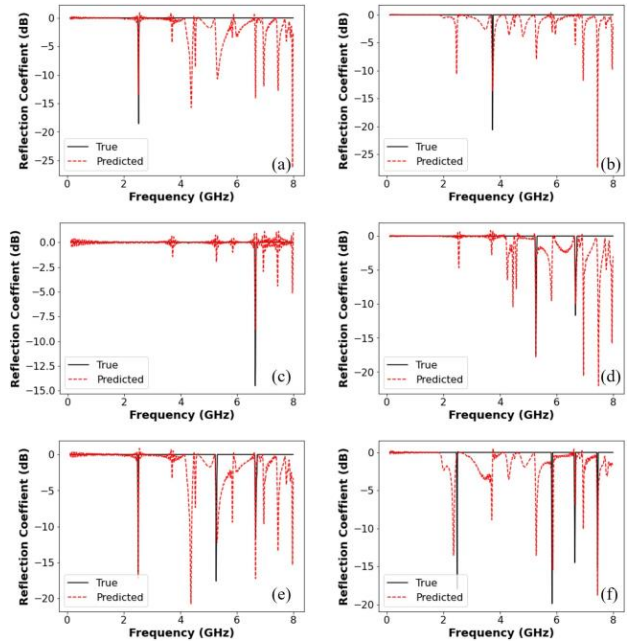

Fig. 7. Simulated (predicted) reflection coefficient curves vs input (true) reflection coefficient at different resonating frequencies

## ACKNOWLEDGEMENT

Authors would like to thank Shri Y Dilip, Director, Aeronautical Development Establishment (ADE), Mr. Manjunath S M, Technology Director and Mr. Diptiman Biswas, Group Director for their support during the research work carried out at ADE, DRDO.